\documentclass[letterpaper, 10 pt, conference]{ieeeconf}

\IEEEoverridecommandlockouts                              % This command is only needed if 
\usepackage[utf8]{inputenc}
\usepackage{comment}
\usepackage{booktabs}

\usepackage{graphicx}
\usepackage{caption}
\usepackage{subcaption}
\usepackage{amsmath}
\usepackage{verbatim}
\usepackage{amsfonts}
\usepackage{multirow}
\usepackage[table,xcdraw]{xcolor}
\usepackage{changepage}
\usepackage{tabularx}
\usepackage{adjustbox}
\usepackage{mathtools}
\usepackage{amssymb}
\usepackage{soul}
\usepackage{color}
\usepackage[noend]{algpseudocode}

\usepackage{color}
\usepackage{tikz}
\usepackage{flushend}
\usetikzlibrary{shapes.geometric, arrows}
\usepackage[linesnumbered,ruled]{algorithm2e}
\usepackage{multirow}
\usepackage{soul,color}
\usepackage{graphicx}
\usepackage[colorlinks]{hyperref}
\usepackage{fancyhdr}
\usepackage{comment}
\usepackage{epsfig}
\usepackage{bm}
\usepackage{float}
\usepackage{cuted}
\usepackage{tcolorbox}
\usepackage{xcolor}
\usepackage{mdframed}
\usepackage{listings}

\allowdisplaybreaks

\makeatletter
\def\algbackskip{\hskip-\ALG@thistlm}
\makeatother

\newcommand{\intersection}{\ensuremath{\operatorname{\cap}}}

\newcommand\oprocendsymbol{\hbox{$\square$}}
\newcommand\oprocend{\relax\ifmmode\else\unskip\hfill\fi\oprocendsymbol}

\usepackage{cite}

\newcommand\red[1]{{\color{red} #1}}
\newcommand\blue[1]{{\color{blue} #1}}

\renewcommand{\theenumi}{(\roman{enumi}}

\mdfdefinestyle{custombox}{%
    backgroundcolor=gray!10, % Define a light gray background
    linewidth=1pt,           % Border thickness
    linecolor=black,         % Border color
    innertopmargin=10pt,     % Top margin inside the box
    innerbottommargin=10pt,  % Bottom margin inside the box
    innerrightmargin=10pt,   % Right margin inside the box
    innerleftmargin=10pt     % Left margin inside the box
}

\makeatletter
\let\NAT@parse\undefined
\makeatother

 \renewcommand{\baselinestretch}{0.99}

\title{\LARGE \bf
Scale-Plan: Scalable Language-Enabled Task Planning for Heterogeneous Multi-Robot Teams}

\author{Anonymous Submission}

\author{
Piyush Gupta\textsuperscript{1*} \hspace{1.2cm}  Sangjae Bae\textsuperscript{1} \hspace{1.2cm} Jiachen Li\textsuperscript{2} \hspace{1.2cm} David Isele\textsuperscript{1} 
\thanks{
\textsuperscript{1} Honda Research Institute, San Jose, CA 95134, USA.}
\thanks{
\textsuperscript{2} Georgia Institute of Technology, Atlanta, GA 30332, USA.}
\thanks{
\textsuperscript{*} Corresponding author \texttt{\{piyush\_gupta\}@honda-ri.com } \
}
}
\begin{document}

\maketitle
\thispagestyle{plain}
\pagestyle{plain}
\pagenumbering{gobble}
%%%%%%%%%%%%%%%%%%%%%%%%%%%%%%%%%%%%%%%%%%%%%%%%%%%%%%%%%%%%%%%%%%%%%%%%%%%%%%%%
\begin{abstract}
Long-horizon task planning for heterogeneous multi-robot systems is essential for deploying collaborative teams in real-world environments; yet, it remains challenging due to the large volume of perceptual information, much of which is irrelevant to task objectives and burdens planning. Traditional symbolic planners rely on manually constructed problem specifications, limiting scalability and adaptability, while recent large language model (LLM)-based approaches often suffer from hallucinations and weak grounding—i.e., poor alignment between generated plans and actual environmental objects and constraints—in object-rich settings. We present Scale-Plan, a scalable LLM-assisted framework that generates compact, task-relevant problem representations from natural language instructions. Given a PDDL domain specification, Scale-Plan constructs an action graph capturing domain structure and uses shallow LLM reasoning to guide a structured graph search that identifies a minimal subset of relevant actions and objects. By filtering irrelevant information prior to planning, Scale-Plan enables efficient decomposition, allocation, and long-horizon plan generation. We evaluate our approach on complex multi-agent tasks and introduce MAT2-THOR, a cleaned benchmark built on AI2-THOR for reliable evaluation of multi-robot planning systems. Scale-Plan outperforms pure LLM and hybrid LLM–PDDL baselines across all metrics, improving scalability and reliability. 
Project website: \url{ https://github.com/honda-research-institute/Scale_Plan} 
%Code and the MAT2-THOR dataset will be made publicly available upon acceptance.

%\todo{Expand abstract to include result highlights.}

\end{abstract}

%%%%%%%%%%%%%%%%%%%%%%%%%%%%%%%%%%%%%%%%%%%%%%%%%%%%%%%%%%%%%%%%%%%%%%%%%%%%%%%%

\section{Introduction}\label{sec:introduction}

Advancements in robotics and artificial intelligence have accelerated progress in long-horizon task planning for heterogeneous multi-robot systems~\cite{liu2025coherent, zhang2024fltrnn}. By leveraging complementary capabilities across diverse robot embodiments, such systems enable teams to accomplish complex missions in real-world settings, including disaster response, warehouse logistics, and large-scale inspection~\cite{9760047, rizk2019cooperative, 9029753}. Realizing this potential at scale requires planning and coordination mechanisms that explicitly account for heterogeneity while ensuring efficient and reliable execution.

\begin{figure}
    \centering
    \includegraphics[width=1\linewidth]{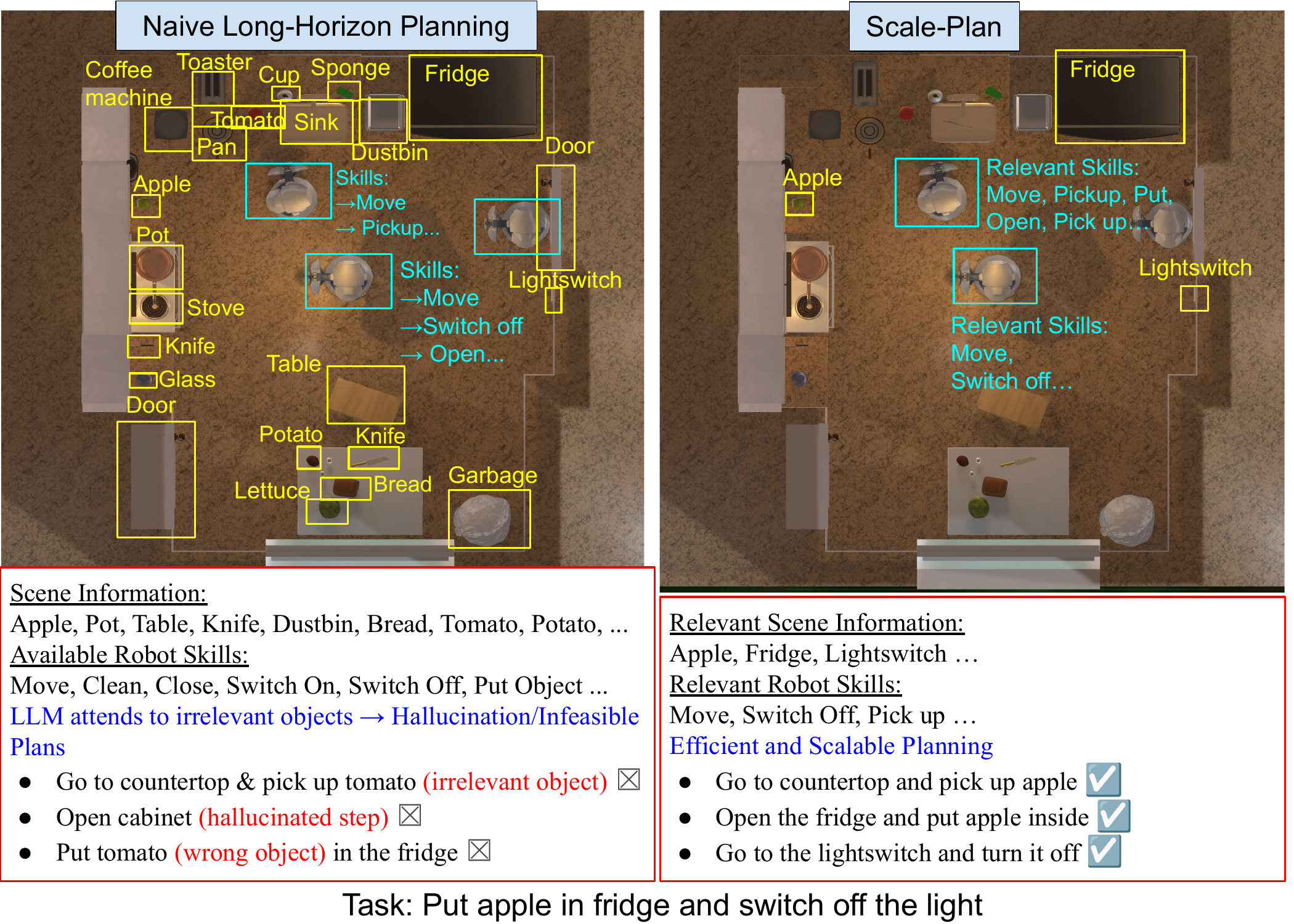}
    \caption{\small Task-relevant environment filtering for scalable planning. Naive long-horizon planning considers all detected objects and robot capabilities, leading to large action spaces and planning errors (left). Scale-Plan extracts only task-relevant scene information and skills (right), significantly reducing combinatorial complexity and enabling efficient multi-robot plan synthesis. } 
    \label{fig:Motivation}
\end{figure}

In real-world environments such as indoor household spaces, long-horizon planning must integrate rich perceptual information. However, much of the available sensory data is irrelevant to the task, creating a major bottleneck for efficient decision-making~\cite{silver2021planning}. For example, as illustrated in Fig.~\ref{fig:Motivation}, when executing a task such as placing the apple in the fridge and turning off the light-switch, only entities directly related to the apple, fridge and the lightswitch are necessary. Including unrelated objects (e.g., tomato, pots, dustbin etc.) unnecessarily enlarges the search space and degrades planning performance~\cite{agia2022taskography}. This challenge is amplified when using large language models (LLMs), which are limited by context length, may hallucinate entities~\cite{gupta2025graph}, and often attend to irrelevant environmental details~\cite{kamath2024llm}. 
Therefore, retaining only task-relevant information is critical for scalable long-horizon planning.

Prior work has sought to address this issue. In~\cite{silver2021planning}, the authors propose a relational learning framework based on graph neural networks~\cite{zhang2019graph} to predict object relevance, but it requires domain-specific training data and offers limited interpretability. Authors of~\cite{agia2022taskography} construct planning domains from dense 3D scene graphs~\cite{armeni20193d} and iteratively sparsify them starting from goal-relevant objects; however, performance is sensitive to initialization and may still retain extraneous entities. Similarly, in~\cite{rana2023sayplan}, the authors leverage LLMs to extract task-relevant subgraphs but restrict the search to node-level expansion or contraction, guided primarily by room semantics. In contrast, we introduce an action-graph representation over PDDL domains to identify task-relevant elements at the domain level.

Traditional symbolic planners~\cite{garrett2018ffrob, lyu2019sdrl} operating over PDDL specifications~\cite{aeronautiques1998pddl} rely on expert-authored domain and problem files. While effective in structured settings, manually constructing detailed problem files for each task limits scalability and adaptability, making such approaches impractical in dynamic real-world environments.

Recent advances in LLMs have motivated their integration into robotic planning~\cite{gupta2025generalized, Kang_2025_CVPR}. Existing approaches either use LLMs directly as planners~\cite{song2023llm, lin2023text2motion} or employ them to generate symbolic problem specifications that are subsequently solved by classical planners~\cite{liu2023llm+, zhang2025lamma}. However, in cluttered environments with limited grounding, LLM-generated representations are often unreliable: they may hallucinate entities, omit constraints, or introduce irrelevant information, resulting in infeasible or inefficient plans~\cite{rana2023sayplan}. Our empirical results highlight both the frequency of such errors and their downstream impact.

To address these limitations, we propose \textbf{Scale-Plan}, a scalable LLM-assisted planning framework for heterogeneous multi-robot systems. Scale-Plan constructs an action graph offline from the PDDL domain, where nodes correspond to parameterized actions and directed edges encode dependencies between action effects and preconditions. This graph captures domain-level action dependencies independent of any instance.

At runtime, given a high-level natural language instruction, Scale-Plan performs shallow LLM-guided reasoning over the action graph to extract minimal task-relevant information. It then decomposes the instruction into subtasks, allocates robots, and synthesizes executable action sequences without explicitly generating an intermediate PDDL problem file. By avoiding exhaustive environmental grounding and manual problem specification, Scale-Plan reduces combinatorial complexity and improves scalability and robustness in long-horizon multi-robot planning.

The main contributions of this work are:
\begin{itemize}
\item We introduce \textbf{Scale-Plan}, a scalable LLM-assisted planning framework that constructs an action graph from PDDL domains to retain task-relevant environmental information, enabling compact problem representations.
\item We develop an LLM-based planning pipeline that decomposes high-level natural language instructions into executable action sequences and performs robot allocation for heterogeneous teams without explicit PDDL problem file generation.
\item We release \textbf{MAT2-THOR}, a cleaned and standardized benchmark derived from MAT-THOR~\cite{zhang2025lamma} for evaluating multi-agent long-horizon planning in AI2-THOR~\cite{kolve2017ai2}.
\end{itemize}

\section{Background and Formulation}\label{sec:background}

A STRIPS-style PDDL planning domain~\cite{aineto2018learning} is defined as a tuple $D=<P,A,\mathcal{T}>$, where $P$ is a finite set of predicates, $A$ is a finite set of parameterized action schemas, and $\mathcal{T}$ defines the transition model through action preconditions and effects.

$Objects$ are typed entities over which predicates and actions are instantiated, representing task-relevant components of the environment (e.g., blocks, robots, tools). Predicates are Boolean relations over $Objects$ that describe properties of the world state (e.g., \textit{at(robot, location), holding(robot, object))}.

Each action schema specifies:
\begin{itemize}
    \item \textbf{Preconditions:} predicates that must hold before execution.
    \item \textbf{Effects:} predicates that become true or false after execution.
\end{itemize}

\begin{figure}[!ht]
\begin{minipage}{0.98\columnwidth}
\centering
\begin{lstlisting}[]
(define (domain simple-pick-place)
 (:requirements :strips :typing)
 (:types robot object location)
 (:predicates (at ?r - robot ?l - location)
              (at-obj ?o - object ?l - location)
              (holding ?r - robot ?o - object))
 (:action pick
  :parameters (?r - robot ?o - object ?l - location)
  :precondition (and (at ?r ?l) (at-obj ?o ?l))
  :effect (and (holding ?r ?o) (not (at-obj ?o ?l)))))
\end{lstlisting}
\caption{Minimal STRIPS-style PDDL domain example used to illustrate predicates, action preconditions, and effects. Scale-Plan constructs its action graph directly from such domain specifications.}
\label{fig:Strips_domain}
\end{minipage}
\end{figure}

\begin{figure}
\begin{minipage}{0.98\columnwidth}
\centering
\begin{lstlisting}[]
(define (problem simple-task)
 (:domain simple-pick-place)
 (:objects
   r1 - robot
   obj1 - object
   locA locB - location)
 (:init
   (at r1 locA)
   (at-obj obj1 locA))
 (:goal
   (holding r1 obj1))
)
\end{lstlisting}
\caption{Example PDDL problem instance. The problem specifies objects, initial state, and goal conditions for a particular task.}
\label{fig:problem_pddl}
\end{minipage}
\end{figure}

\begin{figure*}[!ht]
    \centering
    \includegraphics[width=0.85\linewidth]{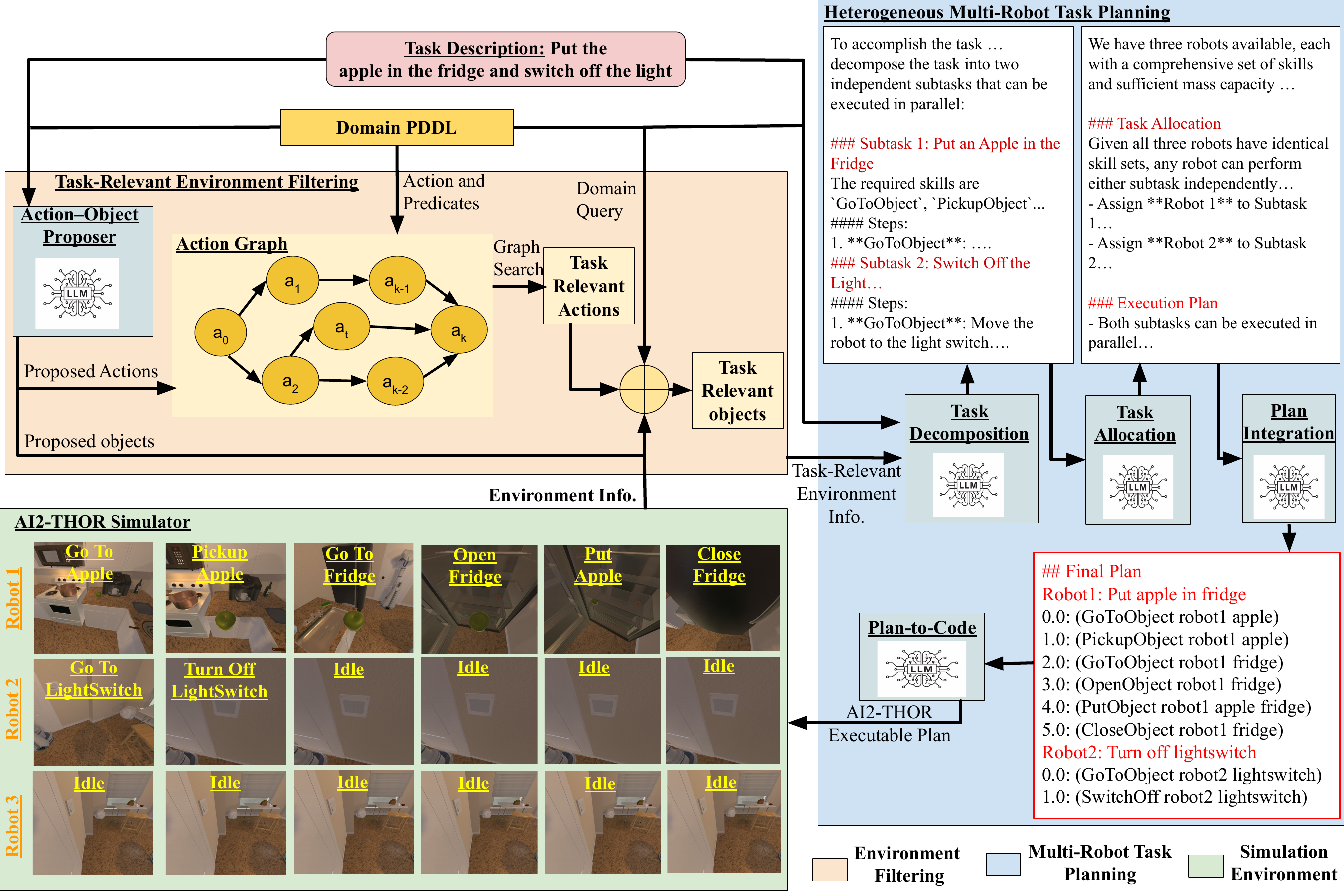}
    \caption{\small Overview of the Scale-Plan framework. Scale-Plan identifies task-relevant environment information by searching an action graph constructed from the PDDL domain. The filtered information is then used by an LLM-based planning pipeline to perform structured task decomposition, task allocation, and plan integration. The resulting high-level plan is translated into an AI2-THOR–executable plan by Plan-to-Code and executed in the simulator.}
    \label{fig:framework}
\end{figure*}
The transition model determines how actions modify the world state by adding or removing grounded predicates. A minimal example domain file is shown in Fig.~\ref{fig:Strips_domain}.

In Scale-Plan, predicates act as the logical interface between actions: an action $a_1$ enables another action $a_2$ if the effects of $a_1$ satisfy the preconditions of $a_2$. 

The planning domain $D$ defines the predicate vocabulary and action schemas independently of any specific instance. A classical planning problem is defined as $<T, D,O,I,G>$, where $T$ is the natural language task description, $O$ is the set of objects, and $I, G \subseteq P$ are grounded predicate sets representing the initial and goal states. 

In the multi robot setting, we define an instance as 
\[\mathcal{M} = <T, R, D, O, I, G>,\] 
where $R=\{r_1, \ldots, r_N \}$ denotes the set of robots. Each robot $r_i$ is associated with an executable action subset $A_i \subseteq A$ determined by its capabilities and constraints, thereby encoding heterogeneity within the shared domain.

The objective is to compute a plan that transforms the state from $I$ to a state satisfying $G$. Scale-Plan extracts task relevant information from $\mathcal{M}$ to construct a reduced domain $D_{fil} = <P, A_{fil}, \mathcal{T}_{fil}>$, where $A_{fil} \subseteq A$ contains only task relevant action schemas and $\mathcal{T}_{fil}$ denotes the transition model restricted to $A_{fil}$. This yields a reduced instance
\[M_{fil} = <T, R, D_{fil}, O_{fil}, I, G>,\]
where $O_{fil} \subseteq O$ contains only task relevant objects. This filtering step reduces combinatorial complexity prior to plan synthesis. The goal is to compute a plan $\Pi_{fil}$ from $\mathcal{M}_{fil}$ such that $\Pi_{fil}$
is also a valid solution to the original instance 
$\mathcal{M}$, i.e., its execution from $I$ satisfies $G$ without violating any action preconditions defined in $D$. 
An example PDDL problem instance corresponding to the full formulation is shown in Fig.~\ref{fig:problem_pddl}.

Many existing approaches generate an intermediate PDDL problem file from $\mathcal{M}$ and rely on a classical planner for solution synthesis~\cite{liu2023llm+, zhang2025lamma}. However, incomplete grounding and the presence of numerous irrelevant objects can lead to failure in problem PDDL construction, resulting in inaccurate or infeasible plans. In contrast, Scale-Plan employs a structured LLM based reasoning pipeline to directly synthesize executable plans from the filtered representation, without explicitly constructing an intermediate PDDL problem file.

\section{Methodology}\label{sec:methodology}

Scale-Plan is designed to address long-horizon task planning for heterogeneous multi-agent systems with scalability as a primary objective. It operates in complex, many-object environments by identifying a minimal subset of robot actions and objects that are relevant to the given task. By restricting reasoning to task-relevant information, Scale-Plan improves planning efficiency, reduces combinatorial complexity, and mitigates LLM hallucinations, leading to more reliable and executable plans.

Fig.~\ref{fig:framework} illustrates the overall architecture of Scale-Plan, which consists of two primary components. The first component performs environment filtering to extract the minimal set of task-relevant actions and objects necessary to solve the original problem. The second component operates on this filtered representation and applies a structured LLM-based pipeline for task decomposition, robot allocation, and plan integration. The design of this planning pipeline is inspired by prior structured LLM planning frameworks~\cite{kannan2024smart, zhang2025lamma}. This separation between information reduction and structured reasoning enables scalable and robust long-horizon planning in heterogeneous multi-robot settings.

\subsection{Action Graph Construction}\label{subsec:action_graph}

% \begin{figure}
%     \centering
%     \includegraphics[width=1\linewidth]{images/Action_graph.png}
%     \caption{\small PlaceHolder for Action Graph}
%     \label{fig:action_graph}
% \end{figure}

\begin{figure}[!ht]
\centering
\begin{subfigure}{0.2\textwidth}
\centering
\includegraphics[width=1\linewidth, height=1.0\linewidth, keepaspectratio]{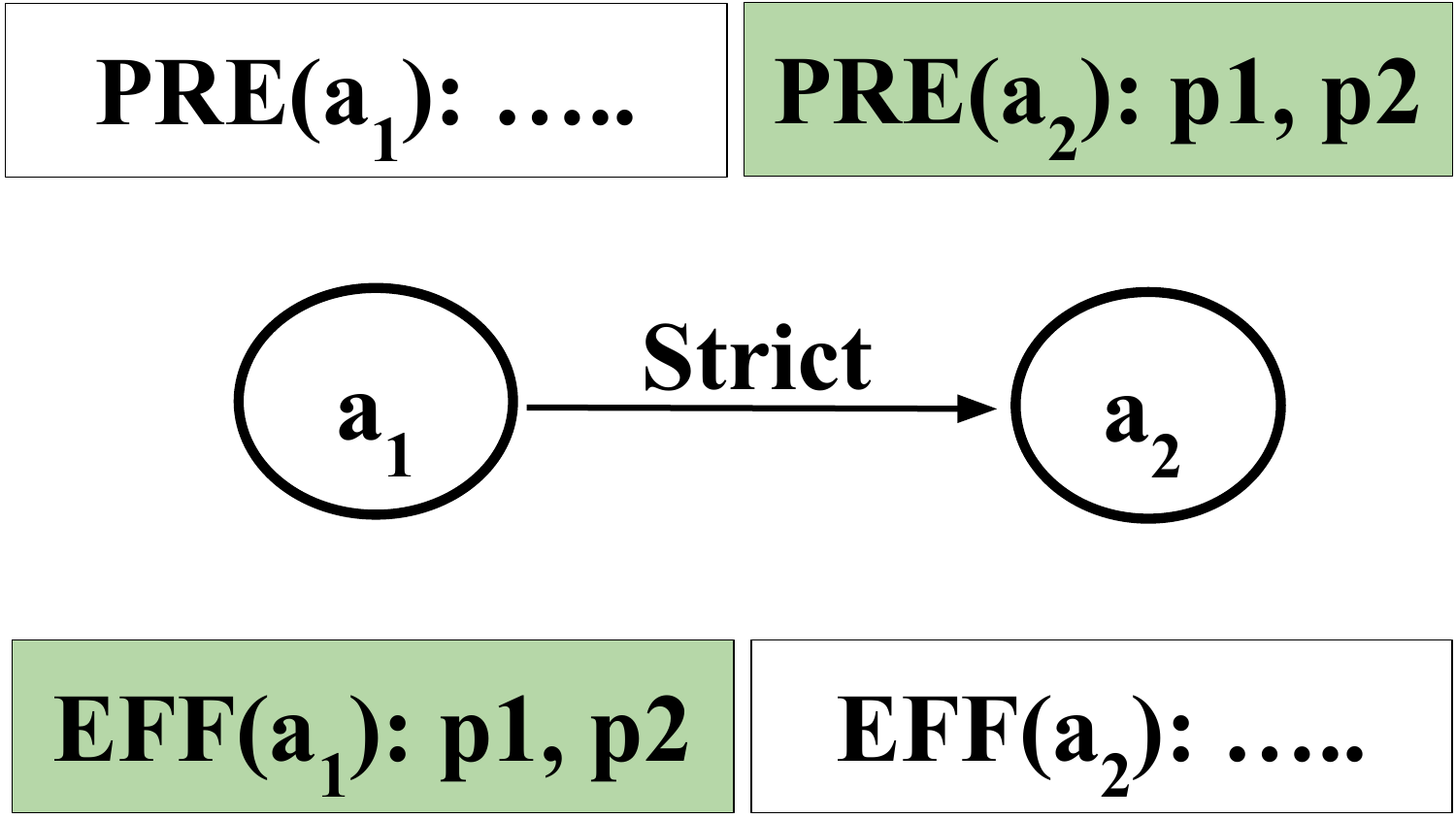}
\caption{Strict Edge}
\label{fig:strict_edge}
\end{subfigure}
\hfill
\begin{subfigure}{0.2\textwidth}
\centering
\includegraphics[width=1\linewidth, height=1.0\linewidth, keepaspectratio]{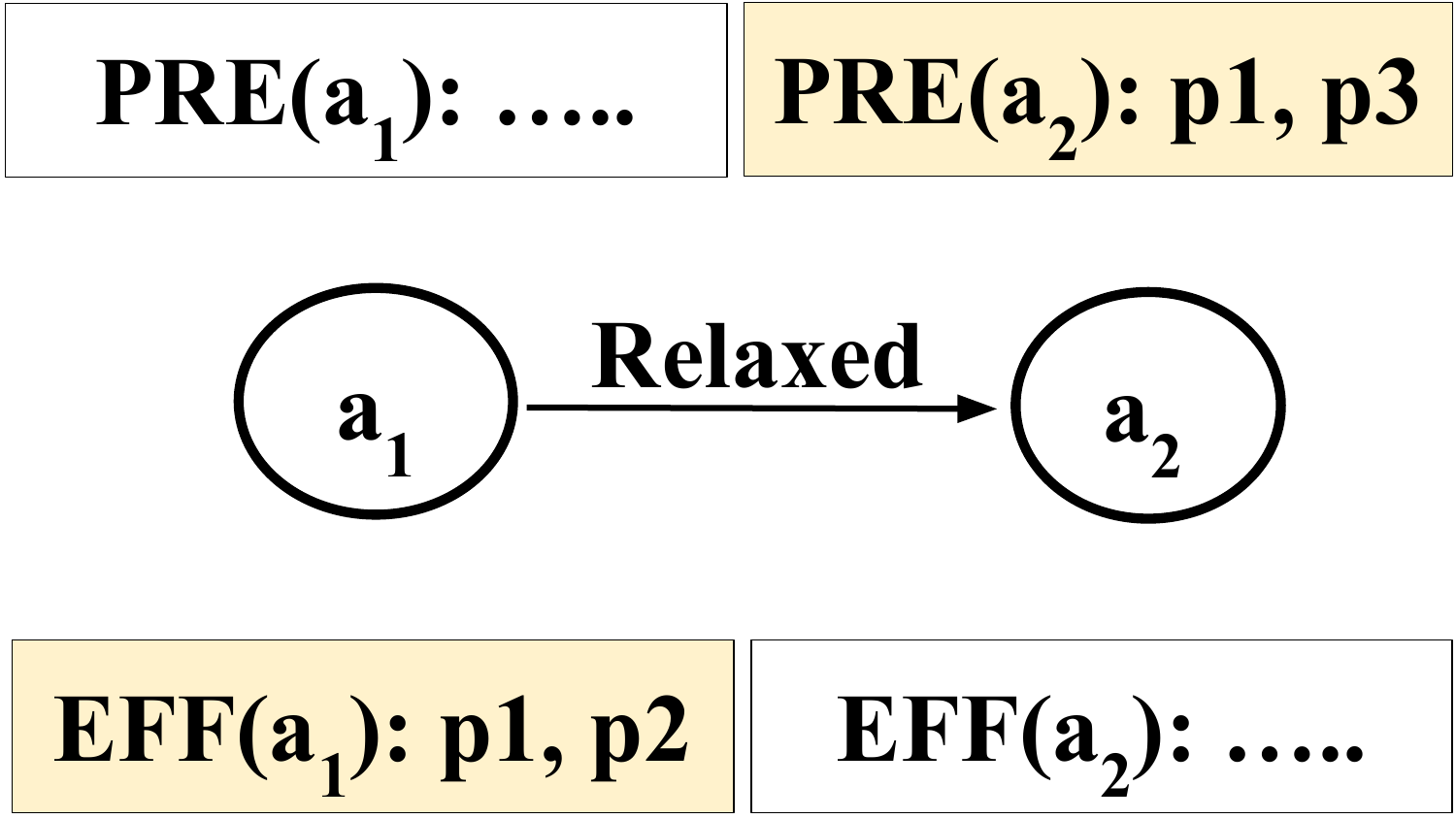}
\caption{Relaxed Edge}
\label{fig:relaxed_edge}
\end{subfigure}

\caption{\small Edge generation in the action graph.
A strict edge $a_1 \rightarrow a_2$ is added if $PRE(a_2) \subseteq EFF(a_1)$. 
A relaxed edge is added when $PRE(a_2) \cap EFF(a_1) \neq \emptyset$ and strict connectivity is absent, ensuring graph connectivity without over-densification.
%\bae{What are PRE and EFF?; my guess is precondition and effect, but please clearly define them.}
%\bae{Also, it is not very clear what ``precondition'' and ``effect'' mean in the first place, and an explanation/example would help.}
}
\label{fig:edge_generation}
\end{figure}

To extract minimal task-relevant information, Scale-Plan constructs an action graph offline from the PDDL domain. In this graph, each node corresponds to a parameterized action schema, and directed edges encode logical dependencies between actions. These dependencies are derived from predicate-level relationships: an action $a_1$ enables another action $a_2$ if the effects of $a_1$ satisfy the preconditions of $a_2$. For each action, we extract its set of preconditions $PRE(\cdot)$ and effects $EFF(\cdot)$. Edges are then generated using two rules as illustrated in Fig.~\ref{fig:edge_generation}.

\begin{enumerate}
    \item Strict Rule: A directed edge $a_1 \rightarrow a_2$ is added if
    \[ PRE(a_2) \subseteq EFF(a_1).\]
This captures a necessary enabling relationship: action $a_1$ provides complete logical support for executing $a_2$.

\item Relaxed Rule: To maintain connectivity, a directed edge  $a_1 \rightarrow a_2$ is added if 
\[PRE(a_2) \intersection EFF(a_1) \neq \emptyset, \]

and either (i) $a_2$ has no incoming strict edges, or (ii) $a_1$ has no outgoing strict edges. This rule captures weaker but meaningful dependencies without over-densifying the graph.
\end{enumerate}

Together, these rules produce a structured directed action graph that preserves essential logical dependencies while avoiding unnecessary connections. Overly strict rules may omit valid dependencies, whereas overly relaxed rules may introduce spurious links. The combined strategy balances structural fidelity and connectivity, reflecting the design intent encoded in the domain.

\textbf{\textit{Illustrative Example:}} Consider actions \textit{PickUp(knife)}, \textit{Slice(tomato)}, and \textit{Place(tomato, plate)}. If \textit{Slice(tomato)} requires the predicate \textit{holding(robot, knife)}, and \textit{PickUp(knife}) produces it, a strict edge is formed from \textit{PickUp} to \textit{Slice}. Similarly, if \textit{Slice} produces \textit{sliced(tomato)} required by \textit{Place}, another strict edge connects them. If no strict predecessor exists but partial predicate overlap occurs, a relaxed edge ensures connectivity. This allows to capture the predicate-level intent of the PDDL domain directly in the form of action graph.

\subsubsection{Task-Relevant Filtering via Graph Search}

At runtime, shallow LLM reasoning first proposes a small set of candidate actions with relevant object parameters conditioned on the task description. These actions are treated as terminal nodes in the action graph. We then perform a backward depth-first search (DFS)~\cite{tarjan1972depth} to identify the minimal subset of predecessor actions required to satisfy their preconditions.

This traversal yields a dependency-consistent action subgraph grounded in the domain structure. Relevant objects are inferred from the parameters of the selected actions, producing a filtered environment containing only the task-relevant actions and objects necessary for planning.

By constructing and searching this domain-level action graph, Scale-Plan systematically reduces combinatorial complexity prior to plan synthesis, while improving the scalability and efficiency of the planner.

\begin{table*}[!ht]
\centering
\resizebox{0.9\textwidth}{!}{
\begin{tabular}{|l|l|l|l|}
\hline
\textbf{\begin{tabular}[c]{@{}l@{}}Task\\ Category\end{tabular}} & \textbf{Task}                                                                                                               & {\color[HTML]{000000} \textbf{MAT-THOR Ground Truth}}                                                                                                                                                 & \textbf{MAT2-THOR Ground Truth}                                                                                                                                                                                                                                              \\ \hline
                                                                 & \begin{tabular}[c]{@{}l@{}}Wash the lettuce and \\ place lettuce on the \\ Countertop\end{tabular}                          & \begin{tabular}[c]{@{}l@{}}{[}\{``name": ``CounterTop", \\ ``contains": {[}``Lettuce"{]}, ``state": None\}{]}\end{tabular}                                                        & \begin{tabular}[c]{@{}l@{}}{[}\{``name": ``CounterTop", ``contains": {[}``Lettuce"{]}, \\ ``state": None\},\\ \blue{ \{``name": ``Lettuce", ``contains": {[}{]},} \\ \blue{``state": ``CLEANED"\}}{]}\end{tabular}                                                           \\ \cline{2-4} 
                                                                 & \begin{tabular}[c]{@{}l@{}}Toast a slice of the \\ breadloaf\end{tabular}                                                   & \begin{tabular}[c]{@{}l@{}}{[}\{``name": ``Bread", ``contains": {[}{]}, \\ ``state": ``COOKED"\}{]}\end{tabular}                                                                    & \begin{tabular}[c]{@{}l@{}}{[}\{``name": ``Bread", ``contains": {[}{]}, ``state": ``COOKED"\}, \\ \blue{\{``name": ``Toaster", ``contains": {[}``Bread"{]}, ``state": ``ON"\},}\\ \blue{ \{``name": ``Bread", ``contains": {[}{]}, ``state": ``SLICED"\} }{]}\end{tabular}            \\ \cline{2-4} 
                                                                 & \begin{tabular}[c]{@{}l@{}}Place the Bar of soap \\ in the sink first and then \\ place dishsponge in the sink\end{tabular} & \begin{tabular}[c]{@{}l@{}}{[} \red{ \{``name": ``Sink", ``contains": {[}``SoapBar",} \\ \red{ ``ScrubBrush"{]} , ``state": None\} } {]}\end{tabular}                                                 & \begin{tabular}[c]{@{}l@{}}{[} 
                                                                 \blue{ \{``name": ``Sink",``contains":{[}``SoapBar",``DishSponge"{]}, } \\ \blue{ ``state": None\}}{]}\end{tabular}                                                                                                                             \\ \cline{2-4} 
\multirow{-4}{*}{\textbf{Simple}}                                & \begin{tabular}[c]{@{}l@{}}Put two vegetables in \\ the fridge parallely\end{tabular}                                       & \begin{tabular}[c]{@{}l@{}}{[}\{``name": ``Fridge", ``contains": {[}``Potato",\\ ``Lettuce"{]}, ``state": None\}{]}\end{tabular}                                                     & \begin{tabular}[c]{@{}l@{}}{[} \blue{ \{``name": ``Fridge", ``contains": {[}``Potato",``Lettuce", } \\ \blue{``Tomato"{]}, ``state": None, ``num\_contains": 2\}}{]}\end{tabular}                                                                                            \\ \hline
                                                                 & \begin{tabular}[c]{@{}l@{}}Wash the lettuce and \\ place lettuce, tomato \\ on the Countertop\end{tabular}                  & \begin{tabular}[c]{@{}l@{}}{[}\{``name": ``CounterTop", \\ ``contains": {[}``Lettuce"{]}, ``state": None\}{]}\end{tabular}                                                        & \begin{tabular}[c]{@{}l@{}}{[}\blue{ \{``name": ``CounterTop", ``contains": {[}``Lettuce", } \\   \blue{ ``Tomato"{]}, ``state": None\},} \\        \blue{ \{``name": ``Lettuce", ``contains": {[}{]},} \\  \blue{ ``state": ``CLEANED"\} }{]}\end{tabular}                                                 \\ \cline{2-4} 
                                                                 & \begin{tabular}[c]{@{}l@{}}Put a plate of slice \\ lettuce into the \\ microwave\end{tabular}                               & \begin{tabular}[c]{@{}l@{}}{[}\{``name": ``Microwave", \\ ``contains": {[}``Plate"{]}, ``state": None\}{]}\end{tabular}                                                            & \begin{tabular}[c]{@{}l@{}}{[}\{``name": ``Microwave", ``contains": {[}``Plate"{]}, ``state": None\},\\  \blue{\{``name": ``Plate", ``contains": {[}``Lettuce"{]}, ``state": None\},}\\  \blue{\{``name": ``Lettuce", ``contains": {[}{]}, ``state": ``SLICED"\} }{]}\end{tabular} \\ \cline{2-4} 
\multirow{-3}{*}{\textbf{Complex}}                               & \begin{tabular}[c]{@{}l@{}}Wash the fork and put it \\ in the bowl, and turn off \\ the light\end{tabular}                  & \begin{tabular}[c]{@{}l@{}}{[}\{``name": ``Bowl", ``contains": {[}``Fork"{]}, \\ ``state": None\}{]}\end{tabular}                                                                   & \begin{tabular}[c]{@{}l@{}}{[}\{``name": ``Bowl", ``contains": {[}``Fork"{]}, ``state": None\},\\  \blue{\{``name": ``Fork", ``contains": {[}{]}, ``state": ``CLEANED"\},}\\  \blue{\{``name": ``LightSwitch", ``contains": {[}{]}, ``state": ``OFF"\}}{]}\end{tabular}            \\ \hline
                                                                 & \begin{tabular}[c]{@{}l@{}}Put a plate of food into \\ the heating appliance\end{tabular}                                   & \begin{tabular}[c]{@{}l@{}}{[}\{``name": ``Microwave", \\ ``contains": {[}``Plate"{]}, ``state": None\}{]}\end{tabular}                                                            & \begin{tabular}[c]{@{}l@{}}{[}\{``name": ``Microwave", ``contains": {[}``Plate"{]}, \\ ``state": None\},\\  \blue{\{``name": ``Plate", ``contains": {[}``Potato", ``Lettuce", }\\  \blue{"Tomato"{]}, ``state": None, ``num\_contains": 1\}}{]}\end{tabular}                  \\ \cline{2-4} 
                                                                 & \begin{tabular}[c]{@{}l@{}}Parallely gather up 3 \\ school supplies on the bed\end{tabular}                                 & \begin{tabular}[c]{@{}l@{}}{[} \red{\{``name": ``Bed", ``contains": {[}``Pen",} \\ \red{"Pensil"{]}, ``state": None\}}{]}\end{tabular}                                                         & \begin{tabular}[c]{@{}l@{}}{[} \blue{\{``name": ``Bed", ``contains": {[}``Pen", ``Pencil", ``Book"{]},} \\ \blue{ ``state": None\}}{]}\end{tabular}                                                                                                                        \\ \cline{2-4} 
\multirow{-3}{*}{\textbf{Vague}}                                 & \begin{tabular}[c]{@{}l@{}}Break all the high tech \\ electronics\end{tabular}                                              & \begin{tabular}[c]{@{}l@{}}{[}\{``name": ``CellPhone", ``contains": {[}{]}, \\ ``state": ``BROKEN"\},\\ \red{\{``name": ``Blinds", ``contains": {[}{]},} \\ \red{``state": ``CLOSED"\}}\end{tabular} & \begin{tabular}[c]{@{}l@{}}{[}\{``name": ``CellPhone", ``contains": {[}{]}, \\ ``state": ``BROKEN"\},\\ \blue{\{``name": ``Laptop", ``contains": {[}{]}, }\\  \blue{``state": ``BROKEN"\}}{]}\end{tabular}                                                                     \\ \hline
\end{tabular}}
\caption{The MAT2-THOR dataset is derived by cleaning and refining the original MAT-THOR dataset. During this process, inconsistent, missing, and erroneous ground-truth conditions (\red{red}) in MAT-THOR were corrected (\blue{blue}). Additionally, we introduce an optional parameter, \text{num\_contains}, which specifies the minimum number of containment conditions that must be satisfied for a task to be considered complete.}
%\di{what is the meaning of the colors, it should be mentioned somewhere?}}
\label{tab:dataset_comparison}
\end{table*}

\subsection{Heterogeneous Multi-Agent Planning}

Using the filtered task relevant representation, the planning pipeline translates high level natural language instructions directly into executable multi robot plans, without constructing an intermediate PDDL problem file. Since problem PDDLs are highly sensitive to object definitions, predicate formats, and goal specifications, even small inconsistencies can lead to infeasible plans. By eliminating explicit problem file generation, which requires precise environmental grounding and strict symbolic formatting, Scale Plan reduces such errors and improves scalability and robustness in complex, object rich environments.

\textbf{\textit{Task Decomposition:}} The planning pipeline first performs task decomposition. Given the natural language instruction, the filtered domain, and the task-relevant object set, the system decomposes the overall objective into a collection of smaller, manageable sub-tasks. For each sub-task, a candidate sequence of actions is generated, ensuring that the decomposition respects domain constraints and logical dependencies.

\textbf{\textit{Task Allocation:}} After task decomposition, the sub-tasks are allocated across heterogeneous robots according to their capabilities and operational constraints. For each sub-task, the required actions are matched to robots whose executable action sets satisfy those requirements. This allocation module also considers opportunities for parallel execution, assigning independent sub-tasks to different robots when feasible in order to reduce overall completion time.

\textbf{\textit{Plan Integration:}} Following allocation, a plan integration module combines the individual sub-task plans into a coherent long-horizon execution strategy. The combiner analyzes inter-task dependencies to determine which actions must be executed sequentially and which can proceed in parallel. By enforcing necessary ordering constraints while enabling concurrency wherever possible, the system maintains logical consistency and improves execution efficiency, particularly in multi-robot scenarios.

\textbf{\textit{Plan-to-Code and AI2-THOR Simulation:}} Finally, the consolidated plan is translated into executable code by a plan-to-code conversion module using structured pattern matching to match the actions executable in AI2-THOR simulator. The resulting code is executed in AI2-THOR.
% The generated code is executed with the AI2-THOR simulator, which visualizes the coordinated behavior of the robots and completes the task execution pipeline.

\section{Results }\label{sec:results}

We now discuss the MAT2-THOR benchmark, experimental setup, relevant baselines, and the evaluation results.

\subsection{Benchmark and Simulation}

We present MAT2-THOR, a cleaned and standardized benchmark derived from MAT-THOR~\cite{zhang2025lamma} for evaluating multi-agent, long-horizon task planning in the AI2-THOR simulator. During our analysis of the original MAT-THOR dataset, we identified several issues that hinder reliable evaluation, including missing or incorrect ground-truth goal conditions, duplicate or redundant tasks, and linguistic inconsistencies in task descriptions. These issues often led to ambiguous success criteria and unreliable plan assessment. Table~\ref{tab:dataset_comparison} presents example tasks from the MAT2-THOR dataset, which was obtained by cleaning and refining the original MAT-THOR dataset. During this process, we identified several issues in the MAT-THOR ground-truth specifications, including missing and erroneous conditions. To resolve ambiguities in containment constraints, MAT2 THOR introduces a variable, \texttt{num\_contains}, requiring that at least \texttt{num\_contains} specified containment conditions be satisfied.

To address these limitations, we systematically curated the dataset by (i) correcting and augmenting ground-truth goal conditions, (ii) removing duplicate or irrelevant tasks, and (iii) refining task descriptions to eliminate language errors while preserving the original intent. The resulting benchmark provides standardized evaluation protocols and consistent success metrics for multi-agent planning.

The final MAT2-THOR benchmark consists of 49 tasks organized into three categories:
\begin{enumerate}
    \item Simple tasks (a total of 25): Simpler tasks with a short-horizon involving one or two subgoals (e.g., “Put the apple in the fridge and switch off the light.”).
    \item Complex tasks (a total of 17): Longer-horizon tasks requiring coordinated execution by heterogeneous robots (e.g., “Put the book in the box, turn on the mobile phone, and clear the desk.”).
    \item Vague tasks (a total of 7): Tasks with underspecified or ambiguous language instructions that require interpretation or implicit goal inference (e.g., “Prepare ingredients for cooking a sandwich tomorrow.”).
\end{enumerate}

MAT2-THOR thus provides a reliable and diverse testbed for evaluating multi-agent planning under varying levels of task complexity and linguistic ambiguity. 

\begin{table*}[!t]
\resizebox{\textwidth}{!}{
\begin{tabular}{|c|c|c|c|c|c|c|c|c|c|c|c|c|c|c|c|}
\hline
\textbf{Method} & \textbf{\begin{tabular}[c]{@{}c@{}}Simple\\ No. Tasks\end{tabular}} & \textbf{\begin{tabular}[c]{@{}c@{}}Simple\\ TCR (\%)\end{tabular}} & \textbf{\begin{tabular}[c]{@{}c@{}}Simple\\ GCR (\%)\end{tabular}} & \textbf{\begin{tabular}[c]{@{}c@{}}Simple\\ ER (\%)\end{tabular}} &  \textbf{\begin{tabular}[c]{@{}c@{}}Complex\\ No. Tasks\end{tabular}} & \textbf{\begin{tabular}[c]{@{}c@{}}Complex\\ TCR (\%)\end{tabular}} & \textbf{\begin{tabular}[c]{@{}c@{}}Complex\\ GCR (\%)\end{tabular}} & \textbf{\begin{tabular}[c]{@{}c@{}}Complex\\ ER (\%)\end{tabular}} &  \textbf{\begin{tabular}[c]{@{}c@{}}Vague\\ No. Tasks\end{tabular}} & \textbf{\begin{tabular}[c]{@{}c@{}}Vague\\ TCR (\%)\end{tabular}} & \textbf{\begin{tabular}[c]{@{}c@{}}Vague\\ GCR (\%)\end{tabular}} & \textbf{\begin{tabular}[c]{@{}c@{}}Vague\\ ER (\%)\end{tabular}} & \textbf{\begin{tabular}[c]{@{}c@{}}Overall\\ TCR (\%)\end{tabular}} & \textbf{\begin{tabular}[c]{@{}c@{}}Overall\\ GCR (\%)\end{tabular}} & \textbf{\begin{tabular}[c]{@{}c@{}}Overall\\ ER (\%)\end{tabular}} \\ \hline
LLM as a Planner & 25 & 44 & 56 & 79 &  17 & 18 & 40 & 69 &  7 & 57 & 57 & 88 & 37 & 51 & 77 \\ \hline
LLM+P (PDDL based) & 25 & 48 & 57 & 74  & 17 & 24 & 39 & 61 & 7 & 43 & 52 & 58 & 39 & 50 & 67  \\ \hline
LaMMA-P (PDDL plan only) & 25 & 36 & 56 & 79 & 17 & 18 & 37 & 66 & 7 & 57 & 64 & 77 & 33 & 51 & 74 \\ \hline
LaMMA-P (LLM corrected) & 25 & 76 & 84 & 91  & 17 & 24 & 56 & 77 & 7 & 43 & 50 & 85 & 53 & 69 & 85 \\ \hline
Scale Plan (Ours) & 25 & \textbf{92} & \textbf{95} & \textbf{98} & 17 & \textbf{59} & \textbf{75} & \textbf{92} & 7 & \textbf{71} & \textbf{71} & \textbf{82} & \textbf{78} & \textbf{85} & \textbf{94}\\ \hline
\end{tabular}}
\caption{Quantitative performance comparison on MAT2-THOR in AI2-THOR. Metrics include Task Completion Rate (TCR), Goal Condition Recall (GCR), and Executability Rate (ER) across simple, complex, and vague task categories.}
\label{tab:performance1}
\end{table*}

\begin{table*}[!t]
\resizebox{\textwidth}{!}{
\begin{tabular}{|c|c|c|c|c|c|c|c|c|c|c|c|c|c|c|c|}
\hline
\textbf{Method} & \textbf{\begin{tabular}[c]{@{}c@{}}Simple\\ No. Tasks\end{tabular}} & \textbf{\begin{tabular}[c]{@{}c@{}}Simple\\ TCR (\%)\end{tabular}} & \textbf{\begin{tabular}[c]{@{}c@{}}Simple\\ GCR (\%)\end{tabular}} & \textbf{\begin{tabular}[c]{@{}c@{}}Simple\\ ER (\%)\end{tabular}} &  \textbf{\begin{tabular}[c]{@{}c@{}}Complex\\ No. Tasks\end{tabular}} & \textbf{\begin{tabular}[c]{@{}c@{}}Complex\\ TCR (\%)\end{tabular}} & \textbf{\begin{tabular}[c]{@{}c@{}}Complex\\ GCR (\%)\end{tabular}} & \textbf{\begin{tabular}[c]{@{}c@{}}Complex\\ ER (\%)\end{tabular}} &  \textbf{\begin{tabular}[c]{@{}c@{}}Vague\\ No. Tasks\end{tabular}} & \textbf{\begin{tabular}[c]{@{}c@{}}Vague\\ TCR (\%)\end{tabular}} & \textbf{\begin{tabular}[c]{@{}c@{}}Vague\\ GCR (\%)\end{tabular}} & \textbf{\begin{tabular}[c]{@{}c@{}}Vague\\ ER (\%)\end{tabular}} & \textbf{\begin{tabular}[c]{@{}c@{}}Overall\\ TCR (\%)\end{tabular}} & \textbf{\begin{tabular}[c]{@{}c@{}}Overall\\ GCR (\%)\end{tabular}} & \textbf{\begin{tabular}[c]{@{}c@{}}Overall\\ ER (\%)\end{tabular}} \\ \hline

No-EF + Joint Task Planning (JTP) & 25 & 80 & 85 & 96 & 17 & 41 & 65 & 90 & 7 & 71 & 71 & 89 & 65 & 76 & 93 \\ \hline 

No-EF & 25 & 80 & 89 & 97  & 17 & 44 & 70 & 91 & 7 & 71 & 71 & \textbf{90} & 66 & 80 & 94 \\ \hline

LLM-Based Shallow Filtering (LLM-SF) & 25 & {88} & {94} & \textbf{99} & 17 & {41} & {70} & {90} & 7 & {57} & {64} & 85 & {67} & {81}  & {94}\\ \hline

ScalePlan (Full Model) & 25 & \textbf{92} & \textbf{95} & {98} & 17 & \textbf{59} & \textbf{75} & \textbf{92} & 7 & \textbf{71} & \textbf{71} & 82 & \textbf{78} & \textbf{85}  & \textbf{94}\\ \hline
\end{tabular}}
\caption{Ablation results comparing performance with and without task-relevant environment filtering. Filtering improves task completion and scalability while maintaining robustness.}
\label{tab:ablation}
\end{table*}

\begin{table}[!t]
\centering
\resizebox{0.8\linewidth}{!}{
\centering
\begin{tabular}{|l|l|l|l|l|}
\hline
\centering
\textbf{Method} &
  \textbf{\begin{tabular}[c]{@{}l@{}}Simple \\ PT (s)\end{tabular}} &
  \textbf{\begin{tabular}[c]{@{}l@{}}Complex \\ PT (s)\end{tabular}} &
  \textbf{\begin{tabular}[c]{@{}l@{}}Vague \\ PT (s)\end{tabular}} &
  \textbf{\begin{tabular}[c]{@{}l@{}}Overall \\ PT (s)\end{tabular}} \\ \hline
\textbf{LLM as a Planner}         & 10.8 & 15.7 & 13.1 & 12.83 \\ \hline
\textbf{LLM+P (PDDL based)}       & 21.4 & 22.2 & 25.9 & 22.32 \\ \hline
\textbf{LaMMA-P (PDDL plan only)} & 52.5 & 62.2 & 64.6 & 57.59 \\ \hline
\textbf{LaMMA-P (LLM corrected)}  & 57.0 & 67.6 & 83.0 & 64.39 \\ \hline
\textbf{Scale Plan (Ours)}        & 53.5 & 72.1 & 69.7 & 62.27 \\ \hline
\end{tabular}}
\caption{Planning Time (PT) comparison across methods, measuring total computation time including LLM reasoning and symbolic planning.}
\label{tab:planning_time}
\end{table}

\subsection{Baselines and Evaluation Metrics}

We compare Scale-Plan against four representative baselines spanning pure LLM-based planning and hybrid LLM–PDDL approaches:

\begin{enumerate}
\item \textbf{LLM as a Planner.}
This baseline directly uses an LLM as a long-horizon planner, mapping natural language task descriptions to executable multi-robot action sequences conditioned on the domain specification~\cite{song2023llm}. No symbolic intermediate representation or external planner is used.

\item \textbf{LLM+P~\cite{liu2023llm+}.}  
This method first prompts the LLM to generate an intermediate PDDL problem file from the natural language task description. The generated problem file, together with the domain file, is then solved using the Fast Downward planner~\cite{helmert2006fast} to produce the final robot plan. This baseline evaluates the effectiveness of LLM-generated symbolic problem formulations without additional structural decomposition.

\item \textbf{LaMMA-P~\cite{zhang2025lamma} (PDDL-only).}  
Similar to LLM+P, LaMMA-P constructs an intermediate PDDL problem file and solves it using Fast Downward. However, it employs a structured and modular pipeline consisting of task decomposition, task allocation, problem generation, PDDL validation, classical planning (Fast Downward), and sub-plan combination. In this baseline, we directly evaluate the plans produced by Fast Downward without any further refinement, isolating the performance of the symbolic planning pipeline.

\item \textbf{LaMMA-P~\cite{zhang2025lamma} (LLM-Corrected).}  
This variant augments the LaMMA-P pipeline by incorporating an LLM-based post-processing stage. After Fast Downward generates a plan from the intermediate PDDL, the LLM refines and corrects the plan using the task decomposition and task allocation outputs as contextual guidance. This step enables recovery from errors arising due to imperfect PDDL generation or symbolic planning failures.

\end{enumerate}

Together, these baselines allow us to evaluate (i) purely neural long-horizon planning, (ii) LLM-assisted symbolic planning, (iii) structured modular symbolic planning, and (iv) hybrid symbolic–neural plan refinement.

\textbf{Evaluation Metrics:} We evaluate Scale-Plan against the baselines using three complementary metrics. (i) \emph{Task Completion Rate (TCR)} measures the percentage of tasks for which all ground-truth goal conditions are satisfied in the final simulator state, reflecting end-to-end task success. (ii) \emph{Goal Condition Recall (GCR)} quantifies the proportion of ground-truth goal conditions achieved in the final state, averaged across tasks, thereby providing a fine-grained measure of partial progress even when full completion is not achieved. (iii) \emph{Executability Rate (ER)} reports the percentage of planned actions that are successfully executed in the simulator, irrespective of their relevance to the task goals, and thus evaluates plan feasibility and robustness to low-level execution failures. 
% (iv)\emph{Planning Time (PT)} denotes the total time required to generate the final multi-robot plan, including all intermediate reasoning and planning steps, capturing computational efficiency. 
Together, these metrics, commonly adopted in prior work~\cite{zhang2025lamma, kannan2024smart}, provide a comprehensive assessment of correctness, robustness, and operational efficiency.

\begin{figure*}[!t]
\centering
\begin{subfigure}{0.8\textwidth}
\centering
\includegraphics[width=1\linewidth, height=1.0\linewidth, keepaspectratio]{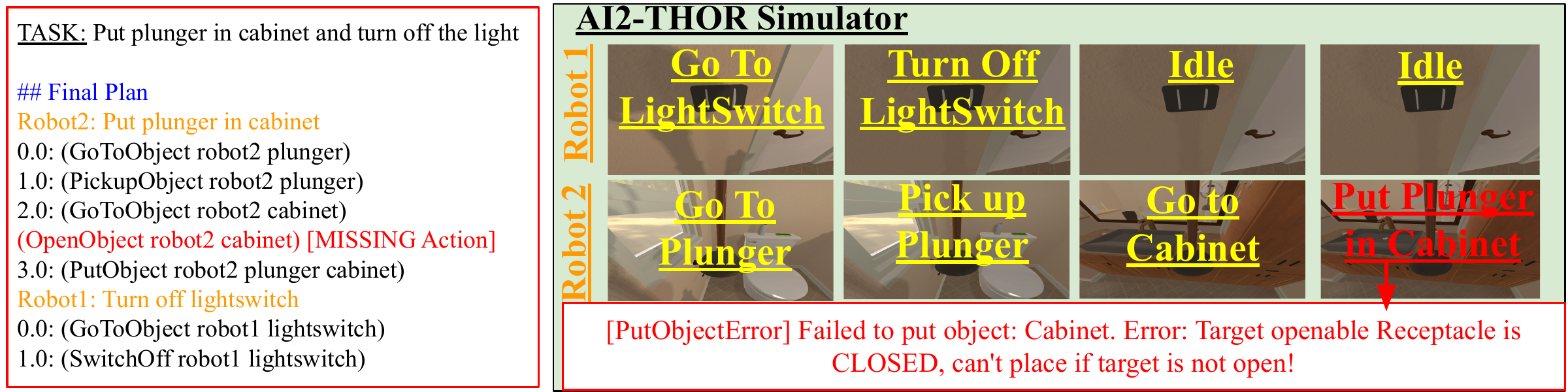}
\caption{Failed to open cabinet before putting plunger.}
\label{fig:failure_ex1}
\end{subfigure}
\hfill
\begin{subfigure}{0.8\textwidth}
\centering
\includegraphics[width=1\linewidth, height=1.0\linewidth, keepaspectratio]{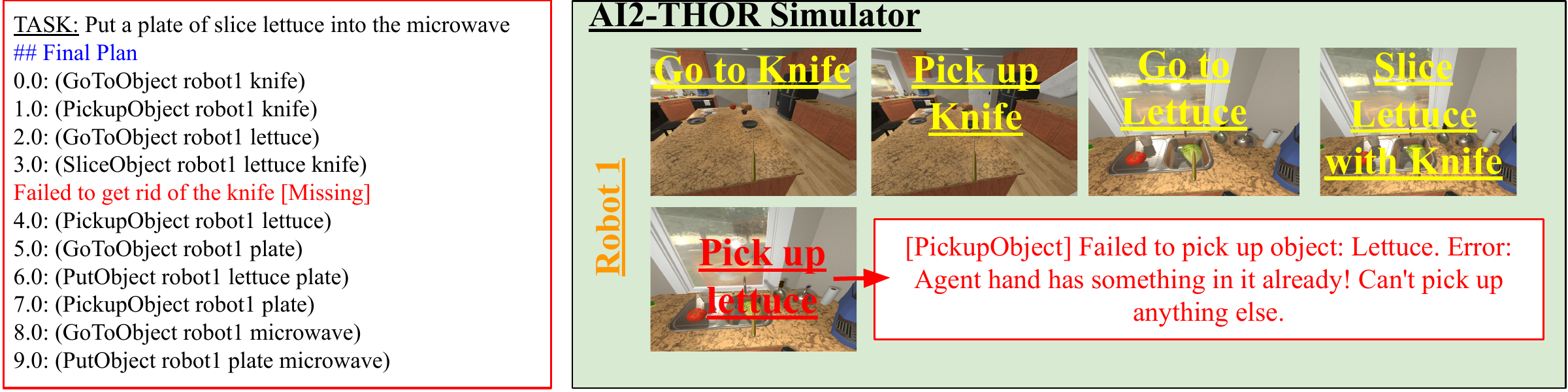}
\caption{Failed to get rid of the knife first before picking up sliced lettuce.}
\label{fig:failure_ex2}
\end{subfigure}

\caption{\small Representative failure cases observed in Scale-Plan. (a) Missing affordance reasoning leads to failure to open a receptacle before placing an object. (b) Missing manipulation constraints cause failure to release a held object before picking up another.}
\label{fig:Quality}
\end{figure*}

\subsection{Quantitative results}

We now present and analyze the simulation results obtained in the AI2-THOR simulator on the MAT2-THOR benchmark. We utilize GPT-5.2~\cite{openai2025gpt5} as the choice of LLM, and evaluate all methods under identical environmental settings and execution protocols to ensure a fair comparison, and report performance across simple, complex, and vague task categories. The results highlight the effectiveness of Scale-Plan in handling long-horizon multi-agent coordination, environmental complexity, and linguistic ambiguity within a realistic embodied simulation environment.

\subsubsection{\textbf{MAT2-THOR evaluation}}

As shown in Table~\ref{tab:performance1}, \textit{Scale-Plan} achieves the best performance across all metrics and task categories. In comparison to the strongest baseline, \textit{LaMMA-P (LLM-corrected)}, our method improves the TCR by 16\% on simple tasks, 35\% on complex tasks, and 28\% on vague tasks, demonstrating consistent and increasing gains as task difficulty and ambiguity increases. Overall, across all task categories in MAT2-THOR, \textit{Scale-Plan} outperforms \textit{LaMMA-P (LLM-corrected)} by 25\% in TCR, 16\% in GCR, and 9\% ER, highlighting its effectiveness in improving both task success and execution reliability.

The \emph{LLM as a Planner} baseline directly uses an LLM to generate robot plans. Without structured symbolic reasoning or explicit grounding, it is prone to hallucinations and logical inconsistencies, particularly in long-horizon settings, often producing infeasible or incomplete multi-agent plans.

The \emph{LLM+P} baseline generates an intermediate PDDL problem file from the task description and uses Fast Downward for planning. However, insufficient grounding and excessive environmental information often yield inconsistent or inaccurate PDDL formulations. Errors in predicates, object definitions, or goals frequently lead to infeasible plans, limiting performance despite the use of a classical planner.

\emph{LaMMA-P (PDDL-only)} introduces a modular pipeline but directly executes plans derived from the intermediate PDDL formulation. Its performance degrades in cluttered environments, where imperfect grounding and noisy state representations propagate into the symbolic problem definition, resulting in inaccurate plans.

\emph{LaMMA-P (LLM-corrected)} is the strongest baseline. Its structured pipeline that includes task decomposition, allocation, problem generation, validation, and sub-plan integration, enables stronger reasoning than LLM+P. An additional LLM-based correction stage repairs some infeasible plans from imperfect PDDL formulations. However, reliance on noisy intermediate representations still limits robustness, especially for complex and vague tasks. In contrast, Scale-Plan tightly integrates structured reasoning with task-relevant environment information, yielding more reliable and scalable multi-agent performance.

%\di{nice discussion of results}

\subsubsection{\textbf{Ablation Study}}
We perform an ablation study to assess the contribution of environment filtering (EF) and the structured planning pipeline in ScalePlan. The full model combines action-graph-based EF with separate task decomposition (TD), task allocation (TA), and plan integration.

We compare against three variants:
\begin{itemize}
    \item No-EF + Joint Task Planning (JTP), which removes filtering and merges TD and TA into a single LLM call;
    \item No-EF, which retains the TD–TA structure but without filtering; and
    \item LLM-Based Shallow Filtering (LLM-SF), which replaces action-graph filtering with shallow LLM filtering.
\end{itemize}

As shown in Table~\ref{tab:ablation}, ScalePlan achieves the highest overall TCR (78\%), outperforming No-EF + JTP by 13\%, No-EF by 12\%, and LLM-SF by 11\%. The improvement is most significant on complex tasks (59\% vs. 41–44\%), highlighting the importance of structured environment filtering for long-horizon reasoning, while maintaining comparable GCR and ER across settings.

\subsubsection{\textbf{Planning Time}}

Table~\ref{tab:planning_time} reports the category-wise and overall \textit{Planning Time (PT)} for all methods, where \textit{PT} denotes the total time required to generate the final multi-robot plan, including all intermediate reasoning and planning steps. This metric, therefore, captures the overall computational efficiency of each approach.

\textit{LLM as a Planner} and \textit{LLM+P} achieve lower \textit{PT}, primarily due to fewer LLM inference calls during planning. In contrast, more structured planning frameworks such as \textit{ScalePlan} and \textit{LaMMA-P} incur higher \textit{PT}, as they require multiple sequential LLM inferences for decomposition, allocation, and coordination. These results highlight a clear trade-off between solution quality and computational efficiency: structured planning strategies improve reasoning and the quality of plans at the cost of increased inference time due to a larger number of LLM calls.

% \todo{Add more ablations}

%  \begin{figure*}[!ht]
%      \centering
%     \includegraphics[width=\linewidth, height=1\linewidth, keepaspectratio]{images/Put_plunger_in_cabinet.pdf}
%     \caption{\small PlaceHolder for Qualitative results}
%     \label{fig:Quality}
% \end{figure*}

\subsection{Failure Cases Analysis}

We analyze common failure modes in ScalePlan. Most failures stem from a small set of recurring issues:

\begin{itemize}
\item The robot navigates to the most likely location of an object rather than its true location, often due to LLM hallucinations in object localization.
\item The robot attempts to pick up a new object without first releasing the object currently held.
\item The robot tries to place an object inside a closed receptacle without opening it beforehand.
\end{itemize}

Fig.~\ref{fig:Quality} illustrates representative examples of these failure cases. In example~\ref{fig:failure_ex1}, while solving the task ``put the plunger in the cabinet and turn off the light”, the generated plan omitted the required action of opening the cabinet before placing the plunger inside. Such errors arise from missing semantic reasoning about object affordances. Incorporating a structured knowledge graph into the planning pipeline that encodes properties such as openable, movable, and cookable could mitigate these issues by enforcing affordance aware action sequencing~\cite{shek2026kglamp}.

In example~\ref{fig:failure_ex2}, during the task ``put a plate of sliced lettuce into the microwave”, the plan failed to release the knife before attempting to pick up the lettuce. This reflects insufficient modeling of manipulation constraints and preconditions. These errors can be mitigated with symbolic constraints and lightweight validation that detect infeasible sequences and trigger corrective fallbacks.

Overall, these failure modes highlight the importance of integrating structured semantic knowledge and constraint-aware validation mechanisms into LLM-based planning frameworks to improve reliability and execution robustness.

\subsection{Limitations and Future Work}\label{sec:Limitations}
% \bae{Consider making it as a subsection of the ``Results" section.}

Our approach has several limitations. First, the absence of explicit environmental grounding, that is, a direct alignment between symbolic or language generated representations such as objects, predicates, and actions and the true simulator state, can lead to LLM hallucinations. This limitation becomes more pronounced in complex, long-horizon tasks that require precise modeling of object states and spatial relationships. Second, in vague or underspecified tasks, the LLM may overlook or incorrectly filter relevant scene objects, resulting in incomplete or suboptimal plans due to ambiguity in the language instructions.

In future work, we plan to incorporate structured knowledge representations, such as knowledge graphs, to provide stronger environmental grounding for LLM-based reasoning. By explicitly aligning language understanding with a structured scene representation, we aim to enable a more reliable intermediate PDDL problem generation stage, thereby improving the accuracy and consistency of generated PDDL formulations. Ensuring well-formed and semantically correct PDDL representations will, in turn, allow classical planners to produce more robust and efficient multi-agent plans. Additionally, developing principled replanning mechanisms to detect and recover from execution failures remains an important direction for future research.
\section{Conclusions}\label{sec:Conclusions}

We presented Scale-Plan, a scalable language-enabled long-horizon planning framework for heterogeneous multi-robot teams operating in object-rich environments. Scale-Plan combines action-graph-based filtering with structured LLM planning to improve scalability and reliability in multi-robot settings.

% Scale-Plan leverages an action graph derived from the PDDL domain to systematically filter task-relevant actions and objects, guided by shallow LLM reasoning. Furthermore, it integrates this structured environment filtering with a planning pipeline that avoids explicit intermediate PDDL problem generation, improving robustness and scalability in long-horizon multi-agent settings.

We also introduced MAT2-THOR, a cleaned benchmark for reliable evaluation in the AI2-THOR simulator. Experimental results demonstrate that Scale-Plan consistently outperforms strong LLM-based and hybrid symbolic baselines. These findings underscore the importance of combining structured domain knowledge with language reasoning for scalable multi-robot task planning.

 % \clearpage
\bibliographystyle{IEEEtran}
\bibliography{ref}

\end{document}